Mini-Project: Applications of Computer Vision in Farming

# A New Vision Algorithm for Detecting the Enzymic Browning Defects in Golden Delicious Apples[1]

Hamid Majidi Balanji *

Email: hamid_majidi2007@yahoo.com

## Abstract

In this work, a new vision algorithm is designed and implemented to extract and identify the surface defects on the Golden Delicious apples caused by the enzymic browning process. 60 Golden Delicious apples were selected for the experiments, of which 30 had enzymic browning defects and the other 30 were sound. Image processing part of the proposed vision algorithm extracted the normalized defective surface area of the apples with the high accuracy of 97.15%. The normalized defective areas of the segmented images were selected as the feature to feed into a threshold-based classifier. The analysis based on the above feature indicated that the images with the normalized defective area less than 0.0065 do not belong to the defective apples; rather, they were extracted as part of the calyx and stem of the healthy apples. The classification accuracy of the classifier applied in this study was 98.5%. The proposed algorithm shows better performance compared to the previously presented methods while needing less computational complexity and hardware cost.

**Keywords:** enzymic browning, vision algorithm, normalized defective areas, Golden Delicious apples.

---

[1] Some part of this paper is represented in 3rd Iranian Agricultural and Environmental Conference, Sanandaj, Iran, 2009.

## 1. Introduction

Enzymic browning is a chemical process involving polyphenol oxidization or other enzymes that create melanin, resulting in a brown color. Enzymic browning is an important color reaction in fruits, vegetables and seafood. Enzymic browning of fruits and vegetables causes heavy economic losses for growers.

The quality of apples strongly depends on the factors such as: shape, color, size, and the absence of surface defects [1]. The quality assessment of fresh fruits has been performed by human inspectors for years. Such methods are not only tedious and time consuming, but also affected by workman interpretation of the surface defects.

Machine vision and artificial intelligence algorithms offer modern and fast solutions to improve post processing operations such as sorting and packaging in agricultural and food industries.

Most research conducted on the methodology of extraction and detection of apple surface defects have suffered from a major disadvantage, that is, the materials and methods required to carry out the experiments have been prohibitively costly.

Unay et al. [1] used a multispectral imaging system for acquiring images from Jonagold apples. They applied several segmentation methods based on thresholding techniques to identify surface-defects. Their acquired images were captured at four wavelengths including 450 *nm*, 500 *nm*, 750 *nm*, and 800 *nm*. They applied three unsupervised classifiers such as *k*-means, competitive neural networks (CNN), and self organizing feature maps (SOM), in addition to ten supervised classifiers such as *k*-nearest neighbor (kNN), linear discriminate classifier (LDC) and others [1].

An experimental machine vision system to identify the putative surface defects of the Pink Lady and Ginger Gold apples has been designed and implemented by Bennedsen et al. [2]. At the core of their system, there has been an imaging system including a multi-spectral camera with two optical filters at 740 *nm* and 940 *nm* and a conveyor belt with rotating cups. The capturing system acquired 6 images of an apple at 60-degree panning angular rotation of the subject per image. Their vision system successfully detected 92% of the surface-defected Pink Lady apples and 90% of the surface defected Ginger Gold apples [2].

Leemans et al. [3] applied hierarchical grading method for detecting the defects of Jonagold apples. They segmented monochrome images based on the differences between the background pixels and the pixels on the object. After feature extraction phase, the objects were classified into clusters by *k*-means algorithm. The fruits were correctly graded with a rate of 73% [3].

In this study, we first capture images from the surface of Golden Delicious apples by an RGB color camera. Then, after preprocessing step, an image processing algorithm is designed by a statistical study to convert the images from True Color RGB images to monochrome images with a special characteristic that makes the resulting monochrome images as the best candidates for analysis based on thresholding techniques. Finally, after selecting the appropriate features as descriptors of the defective and sound apples, the classification task is performed by a threshold-based classifier with a high accuracy.

This paper is organized as follows: In Section 2, we describe materials used for the execution of the experiments and data acquisition, then explain the proposed image segmentation method implemented in this study. In Section 3, we do interactive thresholding directly from histograms, then after image analysis, appropriate feature is extracted and applied to the classifier in order to classify the healthy and defective apples. In Section 4, we provide the results obtained from the experiments and discuss about them. Finally, Section 5 concludes the paper.

# 2. Hardware and Methods

## *2.1 Objectives*

The main focus of this study is to identify and extract the enzymic browning defects. An unavoidable problem encountered in quasi-spherical fruits, is the non-uniform reflectance of light from their surfaces, even when the lighting and imaging system are pre-set to minimize the latter effect, leading to a boundary light reflectance effect that causes segmentation problems at the boundaries of the fruits. One of the goals of this study has been to find a solution for this problem. Fig. 1 shows the block diagram of the proposed vision algorithm. In the following, we will explain each part of the block diagram separately.

## *2.2. Data acquisition*

The experiment site consists of a computer for displaying, storing, and implementing algorithms in order to analyze the digital images of Golden Delicious apples. A digital camera (*Sony Dsc w200 CCD, Japan*) was used for acquiring the images from the surface of apples and white light was used for the scene illumination.

For implementing the vision algorithm, we used MATLAB software. In the following, we explain the preprocessing algorithm steps.

Before the proposed vision algorithm implementation, the dimensions of all acquired images were reduced from 1600×1200 pixels to 1000×750 pixels. The main advantages of size reduction are the reduction in processing time and memory requirements.

All the images acquired for this research were smeared with noise. The latter appeared when segmentation algorithms were applied to their gray level representation. Some low pass filters such as: Ideal, Butterworth and Gaussian were applied to all digital gray level images [4]. Evaluation of their performance with respect to noise suppression showed that the Gaussian low pass filter with the standard deviation of 0.5 best suppresses the noise.

### *2.3. Design of image segmentation algorithm*

Some of the common image segmentation methods like Roberts, Prewitt, Canny, and Sobel edge detection techniques were used in this study for detecting and extracting the apple defects. But, the obtained results were not satisfactory. For instance, by using Robert's edge detection kernel, the whole boundary edge of the apple and also normalized defective area were extracted simultaneously. Even, applying the morphological operators such as erosion could not alleviate the problem. Fig. 2 shows this issue.

Moreover, so far, no published work focusing on the surface defect segmentation using edge-based techniques has been reported.

Thresholding based techniques for defect extraction in agricultural applications are fast, accurate, and reliable. In addition, their hardware and software implementations are simple. One of the most common thresholding techniques is the automatic Otsu thresholding method. However, a thresholding-based segmentation method performs well when the histogram of the gray scale image is bimodal [5]. But as mentioned before, because of the quasi-spherical shape of apple, the latter requirement was not meet.

Fig. 3 demonstrates the histogram of Fig. 2.a. It is observed that the histogram is a multimodal histogram with many valleys and peaks due to the non-uniform light reflectance from the spherical surface of apple. Therefore, applying the automatic Otsu's thresholding technique would result in miss-identification of some healthy parts of the defective apple as enzymic browning defect, culminating error in the classification phase shown. Fig. 4. shows the effect of Otsu's thresholding technique on gray scale image.

The best approach to overcome the above problem is to generate a gray scale image from the True Color RGB image, which has a bimodal histogram, so that the first modal represents the pixels belonging to the defective parts (dark) and the second modal shows the background pixels (bright). Thus, the main focus of this study is to define an RGB to gray scale transformation such that its input being a True Color RGB image and the output to be a corresponding gray scale image with a bimodal histogram which is the best candidate for the automatic Otsu's thresholding technique. The following equation shows the latter hypothetical transformation.

$$f_{\mathrm{gray}} = af_{\mathrm{R}} + bf_{\mathrm{G}} + cf_{\mathrm{B}} \qquad (1)$$

where $f_{\mathrm{gray}}$ is the resultant gray scale image, *a, b, c* are the weights (coefficients) multiplied by the Red ($f_{\mathrm{R}}$), Green ($f_{\mathrm{G}}$), and Blue ($f_{\mathrm{B}}$) components of the *RGB* color space, respectively.

Before computing such a transformation, color sampling operations on the True Color RGB image of the sound and defective parts of a Golden delicious apple was performed. Fig. 5 demonstrates an RGB true-color image of a defective apple. Fig. 6 shows one realization of the color characteristics (color profiles) of the sound and the defective parts of a Golden Delicious apple.

It is observed from Fig. 6 that, the gray level values of Red and Green components of *RGB* color space play an important role in both sound and defective parts of a Golden Delicious apple, while the blue component is almost placed at the bottom of the color profile. The blue light of the spectrum contains a high level of energy in short wave lengths which is the dominant light on the surface of the Earth and present at all image scenes. Therefore, the weight of the Blue component was set to zero and removed from subsequent designs of the transformation functions in this study. Hence, Eq. 1 will be represented as:

$$f_{\mathrm{gray}} = af_{\mathrm{R}} + bf_{\mathrm{G}} \qquad (2)$$

As a statistical study, a permutation operation was performed on the weights of *a* and *b*. We have changed the weights *a* and *b* between 0.1 and 1. The schematic of this operation is shown in Fig. 7.

After implementation of the permutation algorithm, 100 gray scale images for each Golden Delicious apple under experiment were created. Then, all of the resulting images were scrutinized visually and their histograms were computed. Consequently, 39 gray scale images with bimodal histograms as well as visually good appearance were selected as the best candidates for applying the automatic Otsu's thresholding segmentation algorithm in order to extract the defective parts. Table 1 shows the 18 selected weights of *a* and *b*, producing gray scale images with bimodal histograms and the corresponding computed normalized defective surface areas explained later.

After substituting of the 39 weights into equation 2, to generate 39 gray scale images with bimodal histograms, the resulting images were segmented by the automatic Otsu thresholding algorithm, resulting in binary images in which the binary 1's (white) represented the enzymic browning defective parts and the binary 0's (black) showed the background and the healthy parts. Fig. 8.a demonstrates a defective Golden Delicious apple in RGB color space. The computed gray scale image with $a = 0.3$ and $b = 1$, according to Eq. 2, is depicted in Fig. 8.b, and the segmented image obtained from the automatic Otsu's thresholding algorithm is shown in Fig. 8.c.

## 3. Image Analysis, Feature Extraction, and Classification

### *3.1. Image analysis*

It is desired to make sure about the correctness of the extracted defective parts of the Golden Delicious apples by the proposed algorithm and the accuracy of the computed normalized defective area. For the latter purpose, the histograms of all gray scale images were computed, and the best threshold values were chosen from the respective histograms, and verified interactively[2]. Then, we performed thresholding on all gray scale images of the healthy and defective apples. Next, the normalized defective areas (NDA's) of the segmented images were computed. It is worth mentioning that the computed NDA was used as the feature applied to the classifier.

Considering the computed normalized defective area by the proposed technique and the interactive thresholding, to verify the results, we came across the fact that there is a set of Red and Green weights, $a$=0.3 and $b$=1 (according to Eq. 2), that extracts the maximum NDA of defects on apples. Evaluation of the distribution and the arrangement of the Green weights versus those for the Red from Table. 1, as plotted in Fig. 9, has demonstrated that there is no linear relationship between the Green and Red weights. Considering the data arrangement, it has been further observed from Fig. 9 that these coefficients form a triangular configuration.

### *3.2. Feature Extraction*

As a new study, we computed the gravity center of the triangular shape, as 0.7641 for $a$ (the weight multiplied by the red component weight) and 0.7436 for $b$ (the weight multiplied by green component), respectively. Then all *RGB True Color* images of Golden Delicious apples were converted to the gray scale images with a prominent characteristic of having bimodal histograms. Next, after segmenting the resulting gray scale images by the automatic Otsu's thresholding technique, the normalized defective areas of the segmented images were computed. The normalized defective area (NDA) of the segmented image is defined as the sum of number of binary 1 pixels divided by the total number of pixels (binary 1's and 0's) in the segmented image. Tables 2 and 3 show the NDA of the segmented images for defective and sound apples with the weights ($a$, $b$) = (0.3, 1) and ($a$, $b$) = (0.7641, 0.7436).

---

[2] This was done by the facility supported by IPT software enabling a user to surf an image and observe the gray level values of its pixels.

Scrutinizing the above (now binarized) segmented images indicates that the normalized defective area of segmented image is the best descriptors of an apple's putative defect. Furthermore, this feature has been utilized as the input to a threshold-based classifier used in this study which is presented in Tables 2 and 3.

### ***3.3. Threshold Based Classifier***

Considering the normalized defective area of binary segmented images as descriptor to identify whether a given apple's image is healthy or defective, we got to this conclusion that there is a normalized defective area threshold equal to 0.0065 such that if the computed NDA of the segmented image is less than that, then the apple belongs to healthy class. The binary 1's pixels in these kinds of images are representatives of calyx and stem, not defective parts. While if the NDA of the segmented image for a given apple's image is greater than 0.0065, then it is sorted as a defective class. That is:

```
if NDA <= 0.0065
   Given apple is healthy.
else
   Given apple is defective.
```

Fig. 10 indicates the schematic of the threshold-based classifier used in this study as a simple classification tool.

## 4. Results and Discussion

The proposed machine vision algorithm was applied to a batch of 60 Golden Delicious apples, where 30 apples had the enzymic browning defect and the remainders were sound and standard.

Quasi-spherical shapes of apples caused the non-uniform reflectance light from their surface. The latter caused the gray level values of pixels representing an apple surface to fluctuate significantly. Thus, applying segmentation algorithms like edge detection based on thresholding were inefficient. Consequently, the problem of defect identification becomes, to some extent, insurmountable.

We have attempted to find the best segmentation method for extracting defects, based on an adaptive thresholding technique. It is noteworthy that due to non-uniform light reflectance, causing multimodal histograms, common thresholding techniques do not render acceptable performance.

Hence, we developed a vision algorithm based on a transformation whose input is a True Color RGB image and its output, a gray level image with a bimodal histogram. The resulting gray scale image would be a good candidate for applying the automatic Otsu thresholding algorithm. The proposed algorithm is capable of segmenting defects with high accuracy. Interactive thresholding based on choosing an appropriate threshold from the histograms was applied for verification to confirm the accuracy of the proposed vision algorithm. Considering the data in Table 4, we have found that the total normalized defective area computed from the weights ($a$=0.3, $b$=1) is 0.8754 and the total normalized defective area computed by the interactive thresholding technique, as testifier, is 0.5867. The error $e$ is calculated as:

$$e = (NDA_{\mathrm{ts}} - NDA_{\mathrm{hb}})/NDA_{\mathrm{hb}} \times 100 \qquad (3)$$

where $NDA_{ts}$ is the total normalized defective area obtained by the specified weights and $NDA_{hb}$ is the total defective normalized area computed by the interactive thresholding method.

In this case, the accuracy of the enzymic browning defect extraction for the weigh pair ($a$ =0.3, $b$ = 1) would be 49.94%.

In the same manner, the total normalized defective area obtained by using the weight pair ($a$= 0.7641, $b$=0.7436), is 0.6014 and the accuracy of the defect extraction would be computed as 97.51%. So, we choose the weights 0.7641 and 0.7436 for $a$ and $b$, respectively. Therefore, the proposed transformation for this study has been the following:

$$f_{gray} = 0.7641\, f_R + 0.7436\, f_G \qquad (4)$$

Noting Fig. 11 and considering the data of Table 3, a cardinal point has been deduced that all segmented images with the normalized defective areas (NDA's) less than 0.0065, belong to the stem and the calyx of the good apples rather than the defective ones.

The classification accuracy of the threshold-based classifier used in this study is 98.5%, We can conclude that NDA is appropriate feature to define surface defects, and threshold-based classifier has high capability of classification of apples as defective or sound.

The proposed vision algorithm has two distinct advantages over the methods implemented by Bennedsen et al. [2] and Leemans et al. [3]. First, it benefits from a high accuracy of 97.51% in image processing phase and about 98.5% in classification, as compared to 92% recognition for Pink Lady and 90% for Ginger Gold apples reported by Bennedsen et al [2], and 73% reported by Leemans et al [3]. Furthermore, our proposed vision algorithm utilizes low-cost equipment to accomplish the said tasks, while Bennedsen [2], Leemans [3], and Unay et al [1] achieved their results with higher cost equipments such as multispectral cameras and conveyor to conduct their experiments. Although, the approach and techniques in this study use lower cost equipments, the performance in identifying and classification of the surface defects is superior to those reported in the above-mentioned papers.

## 5. Conclusions

In this research, a machine vision algorithm has been developed. First, we proposed *RGB* to gray scale transformation which yielded gray level images with bimodal histograms using Eq. 4. The latter made the application of automatic thresholding approach to image segmentation such as Otsu's adaptive thresholding technique to be possible with a high accuracy. As a result, we obtained 97.51% accuracy for extraction of the apple's defective parts. Then, the normalized defective area (NDA) of the segmented image was used as the feature. The results indicate that segmented images with the normalized defective area (NDA) less than 0.0065 do not belong to defective apples, whereas they are part of the calyx and stem of the healthy apples. Consequently, the NDA was applied to a threshold-based classifier to identify the healthy and defect images. The classification accuracy of the classifier is 98.5%.

Comparing the results obtained from the proposed algorithm with those of the other methods published before, it is concluded that the new method achieves better performance with less cost and computational complexity.

## Acknowledgment

The author would like to thank Prof. William Pratt and Dr. Behnam Ashjari for their helpful comments regarding this paper.

Table 1. Several values of the coefficients $a$ and $b^*$ used in Eqs. 1, 2, 3, and 4, and the respective computed NDA.

| $a$ | $b$ | Normalized Defective Area (NDA) |
|---|---|---|
| 0.3 | 1 | 0.0432 |
| 0.4 | 0.9 | 0.0418 |
| 0.5 | 0.7 | 0.0410 |
| 0.5 | 0.8 | 0.0406 |
| 0.6 | 0.7 | 0.0398 |
| 0.4 | 1 | 0.0394 |
| 0.5 | 0.9 | 0.0385 |
| 1 | 0.2 | 0.0382 |
| 0.8 | 0.5 | 0.0378 |
| 0.6 | 0.8 | 0.0378 |
| 0.7 | 0.7 | 0.0368 |
| 0.5 | 1 | 0.0367 |
| 1 | 0.3 | 0.0361 |
| 0.8 | 0.6 | 0.0359 |
| 0.6 | 0.9 | 0.0359 |
| 0.9 | 0.5 | 0.0351 |
| 1 | 0.4 | 0.0343 |

Table 2. The normalized areas of the segmented images for $a$ = 0.3 and $b$ = 1 for defective and sound apples

| **Apple No.** | **Defective apples** | **Sound apples** |
|---|---|---|
| | **Normalized Defective Area (NDA)** | **Normalized Extracted Area** |
| 1 | 0.031 | $2.6\times10^{-6}$ |
| 2 | 0.1097 | 0.011 |
| 3 | 0.0625 | 0.0238 |
| 4 | 0.0672 | 0.0302 |
| 5 | 0.0303 | $2.55\times10^{-5}$ |
| 6 | 0.0311 | $2.4\times10^{-4}$ |
| 7 | 0.0432 | $1.21\times10^{-4}$ |
| 8 | 0.0591 | $3.14\times10^{-4}$ |
| 9 | 0.0662 | $5.63\times10^{-4}$ |
| 10 | 0.0378 | $4.39\times10^{-4}$ |
| 11 | 0.0423 | $4.88\times10^{-4}$ |
| 12 | 0.0465 | $3.59\times10^{-4}$ |
| 13 | 0.0508 | $2.83\times10^{-4}$ |
| 14 | 0.0467 | $1.16\times10^{-4}$ |
| 15 | 0.0701 | $2.11\times10^{-4}$ |
| 16 | 0.0305 | $2.27\times10^{-4}$ |
| 17 | 0.0487 | $1.86\times10^{-4}$ |

Table 3. The normalized areas of the segmented images for $a$= 0.7641 and $b$= 0.7436 for defective and sound apples.

| Apple No. | Defective apples | Sound apples |
|---|---|---|
| | **Normalized Defective Area (NDA)** | **Normalized Extracted Area** |
| 1 | 0.0246 | 0 |
| 2 | 0.0836 | $4.72\times10^{-4}$ |
| 3 | 0.0530 | 0.0065 |
| 4 | 0.0119 | 0.0095 |
| 5 | 0.0224 | 0 |
| 6 | 0.0344 | $1.56\times10^{-6}$ |
| 7 | 0.0413 | $5.63\times10^{-5}$ |
| 8 | 0.0388 | $1.74\times10^{-4}$ |
| 9 | 0.0224 | $4.01\times10^{-4}$ |
| 10 | 0.0236 | $2.19\times10^{-4}$ |
| 11 | 0.0316 | $1.74\times10^{-4}$ |
| 12 | 0.0401 | $1.59\times10^{-4}$ |
| 13 | 0.0181 | $2.18\times10^{-4}$ |
| 14 | 0.0508 | $3.07\times10^{-5}$ |
| 15 | 0.0275 | $1.27\times10^{-4}$ |
| 16 | 0.0103 | $7.66\times10^{-5}$ |
| 17 | 0.0400 | $9.69\times10^{-5}$ |

Table 4. Errors obtained by the proposed thresholding technique and by interactive thresholding used to evaluate different values of *a* and *b*.

| **Defect apples** | **Weights** | | **Weights** | |
|---|---|---|---|---|
| | ***a*** | ***b*** | ***a*** | ***b*** |
| | **0.3** | **1** | **0.7641** | **0.7436** |
| Total normalized defective Area* | 0.8754 | | 0.6014 | |
| Total normalized defective Area♦ | 0.5867 | | 0.5867 | |
| Error (%) | 50.06% | | 2.49% | |

*Total normalized defective area computed by the proposed thresholding method in this research.

♦ Total normalized defective area computed by the interactive histogram-based thresholding to verify the latter.

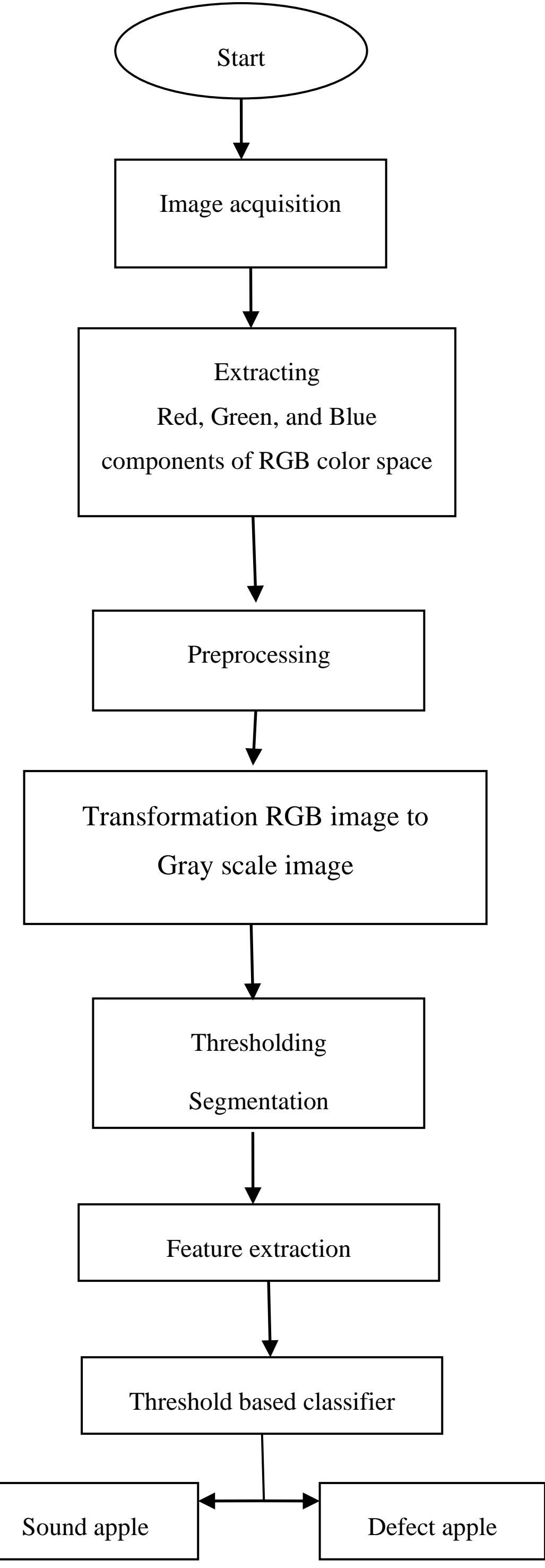

Fig. 1. Block Diagram of the proposed vision algorithm.

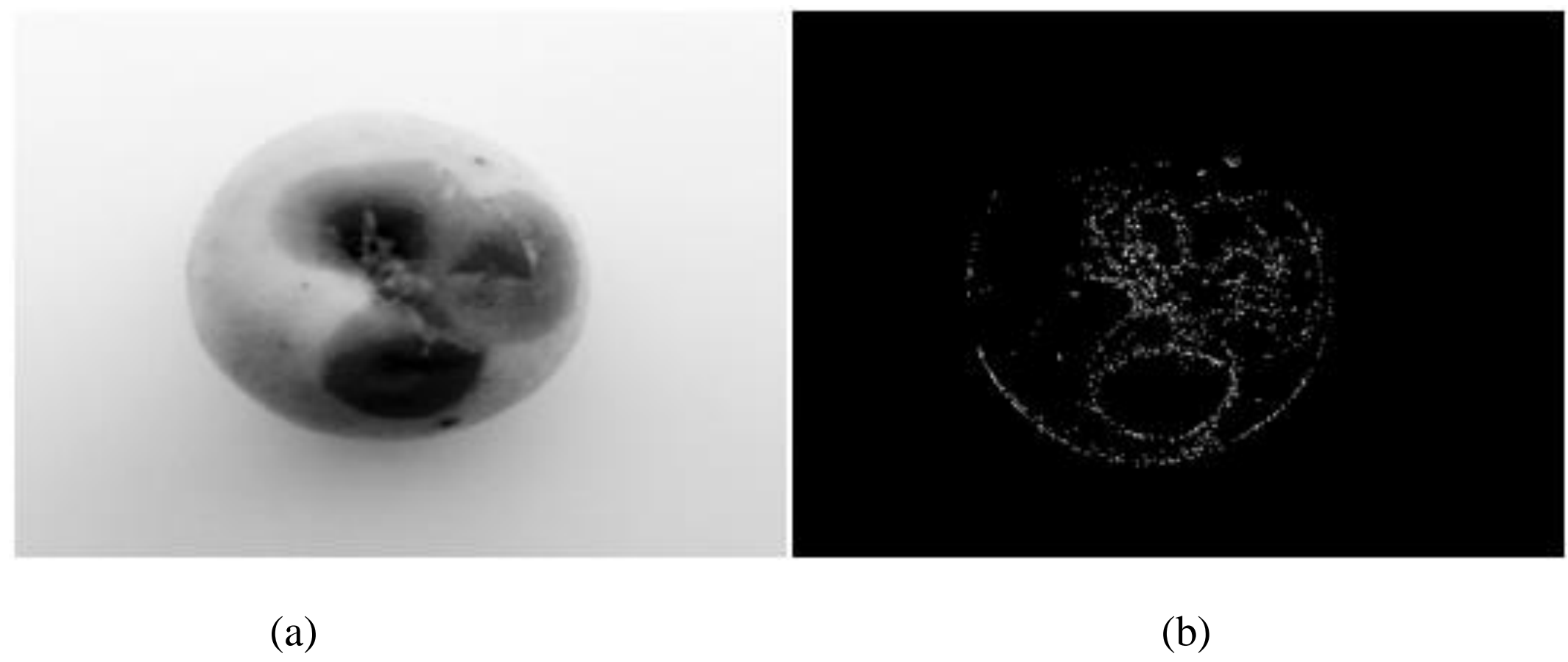

Fig. 2. a) Filtered gray scale image b) Edge detection by Robert’s mask.

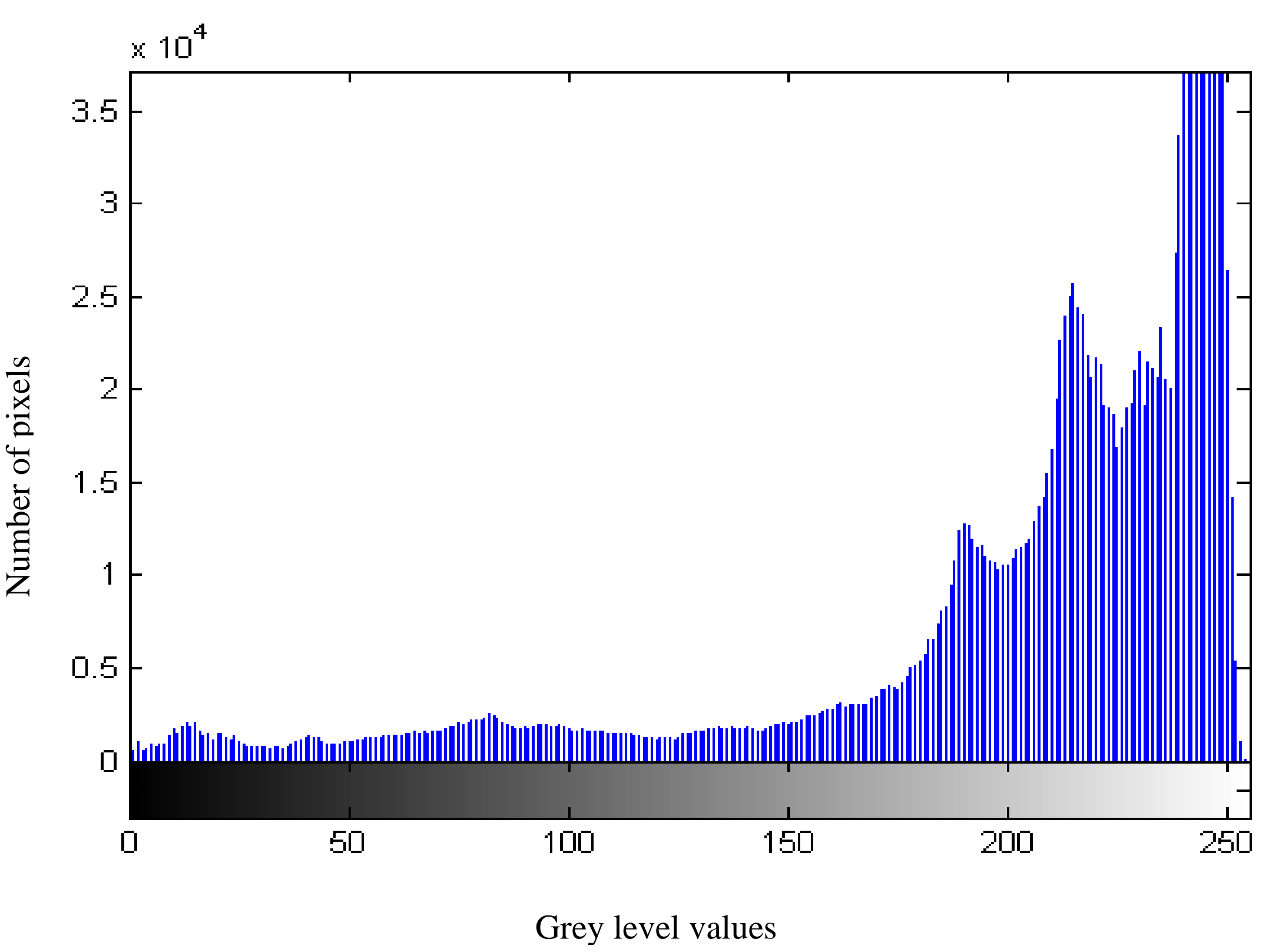

Fig. 3. Histogram of Fig. 2.a.

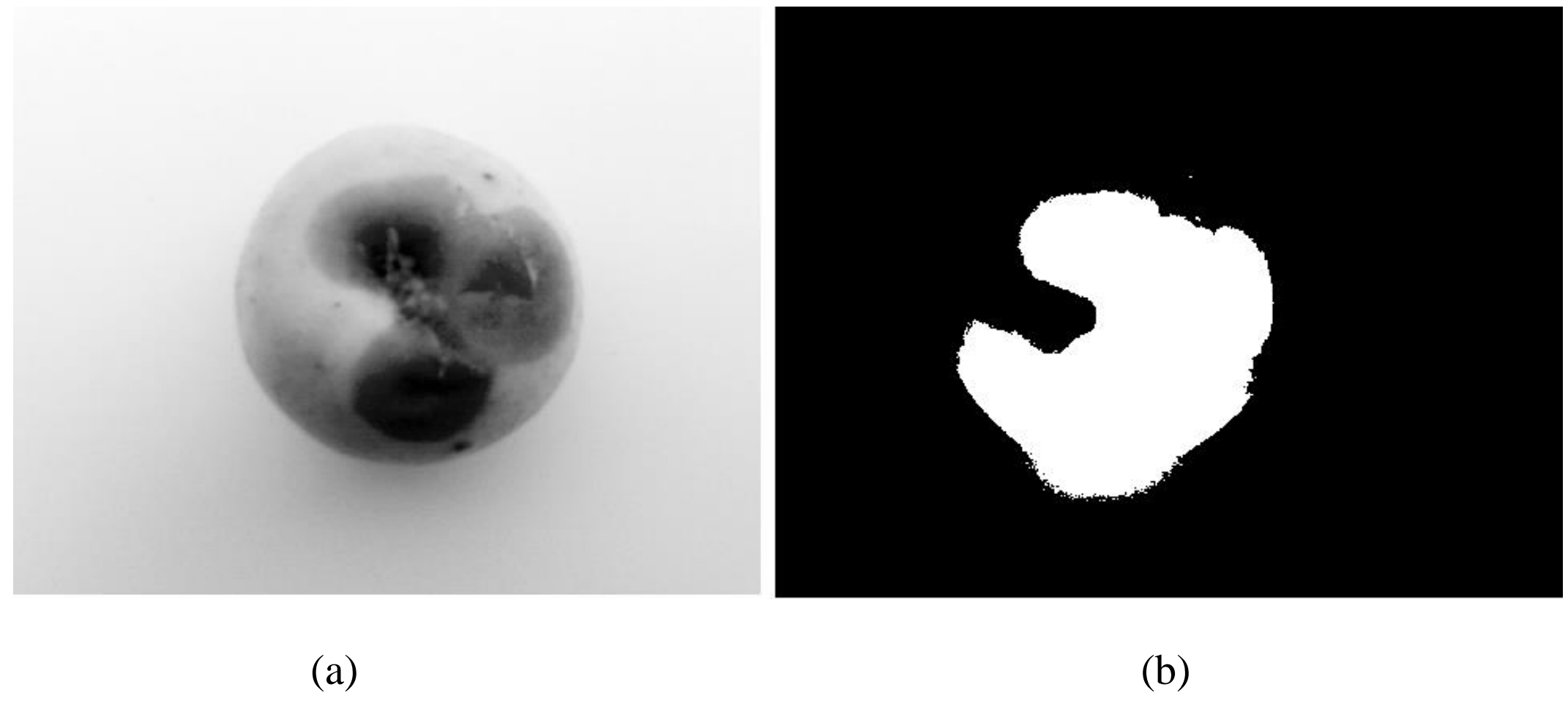

(a) (b)

Fig. 4. a) Grey-scale image

b) Result of Otsu's thresholding technique based on the histogram of Fig. 3.

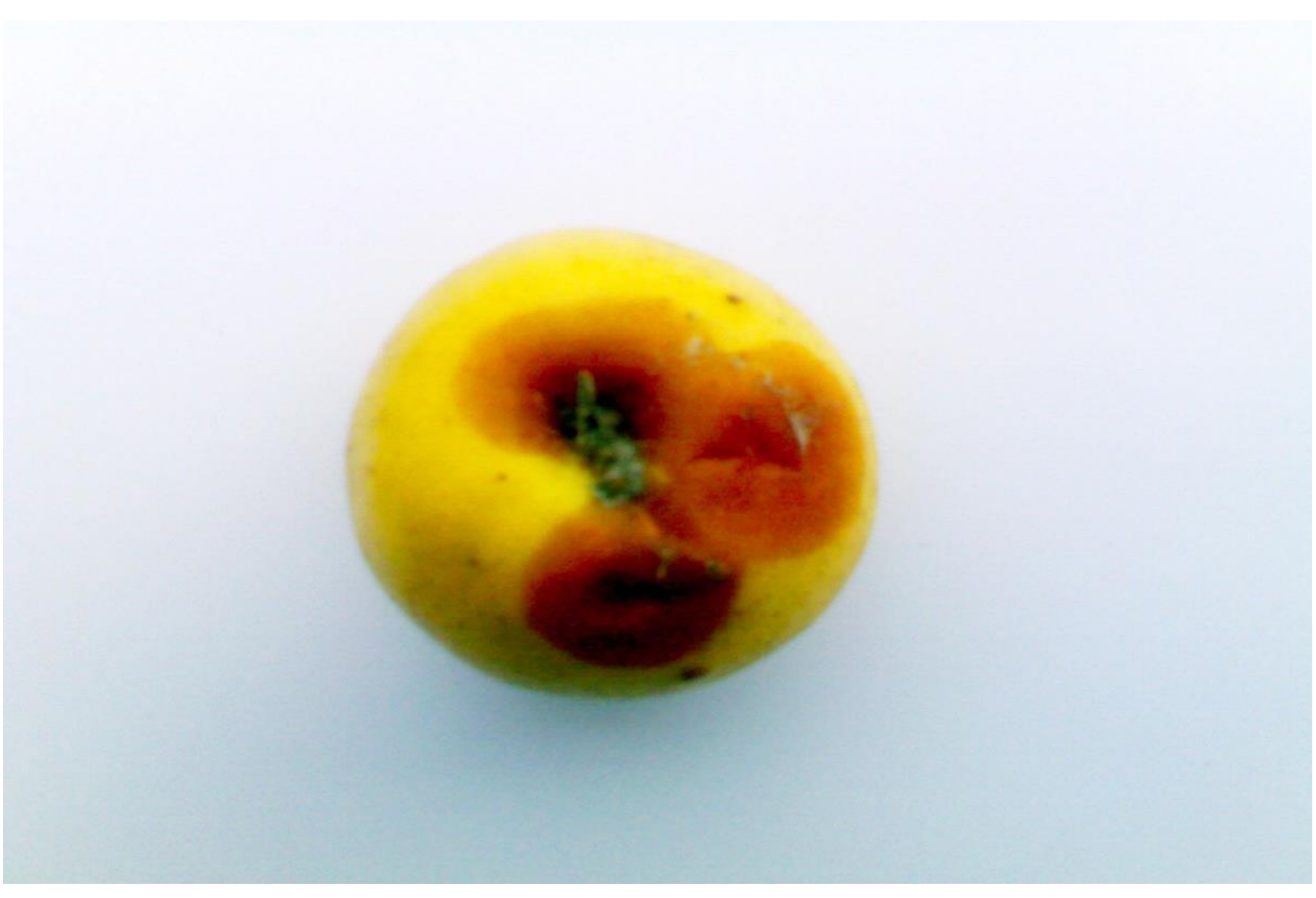

Fig. 5. An RGB true-color image of a defective apple.

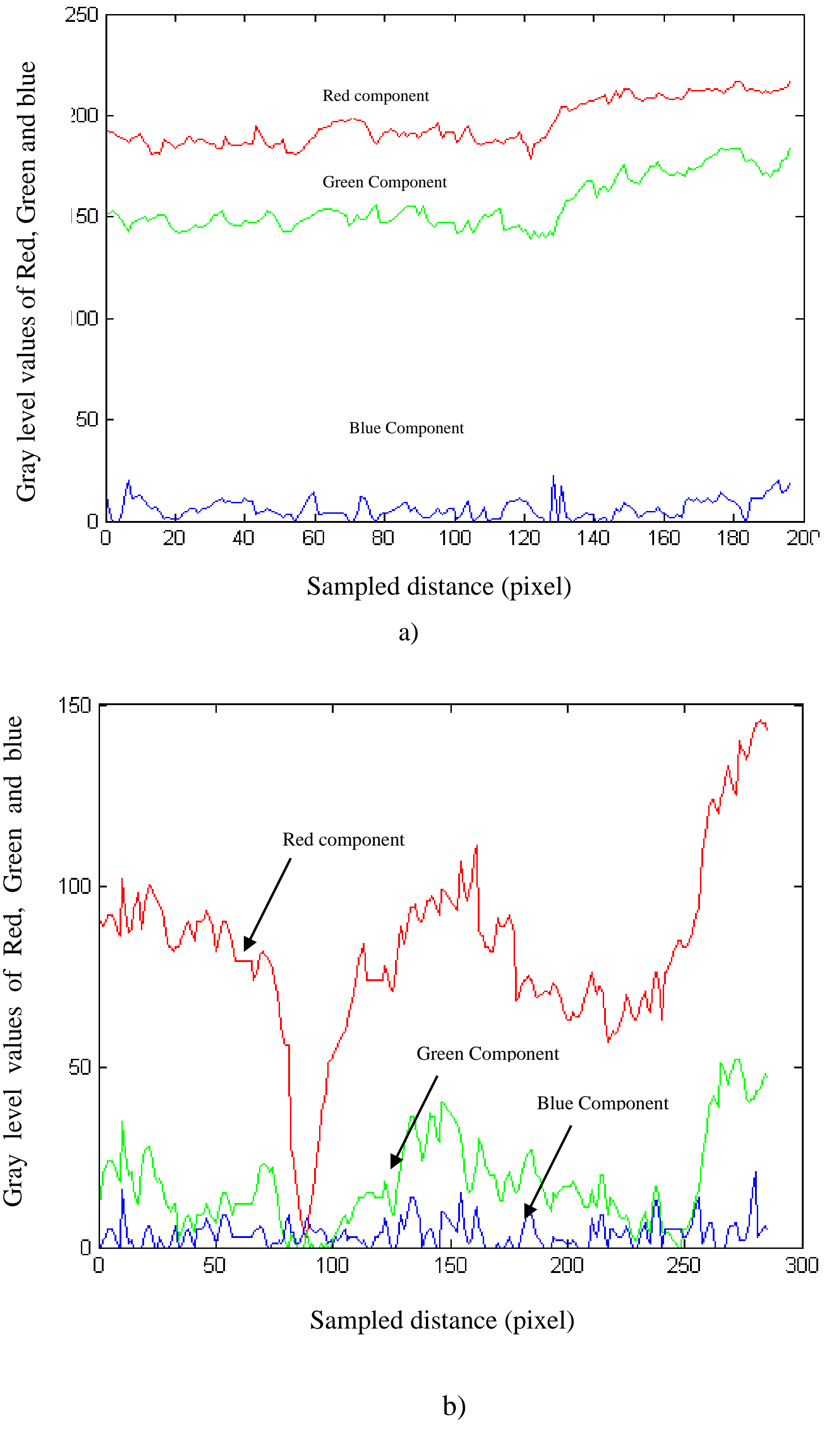

Fig. 6. The RGB components of the golden delicious apple shown in Fig. 5.

a) Sound part b) Defective part.

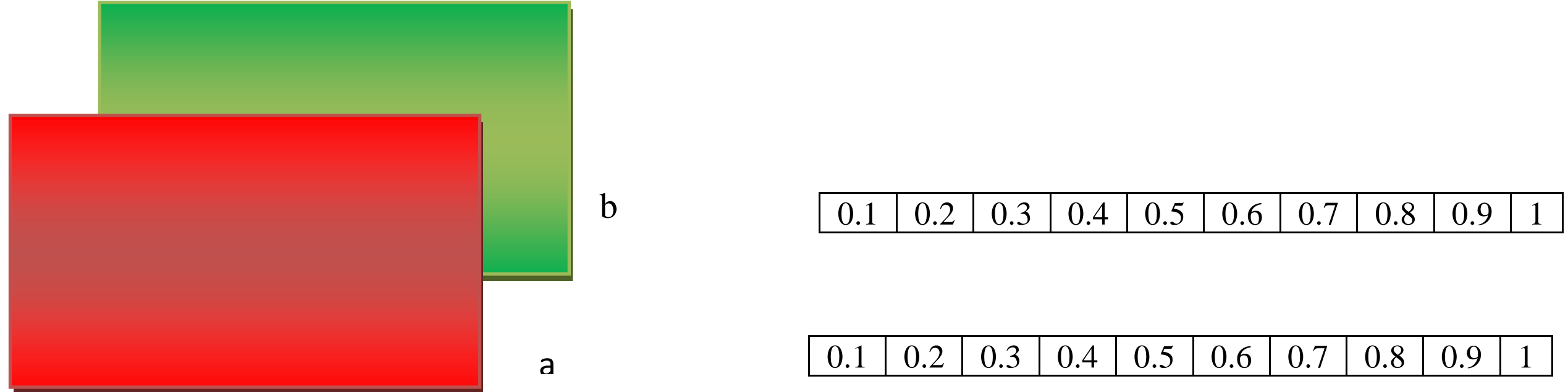

Fig. 7. Schematic of permutation operation on weights of red and green components.

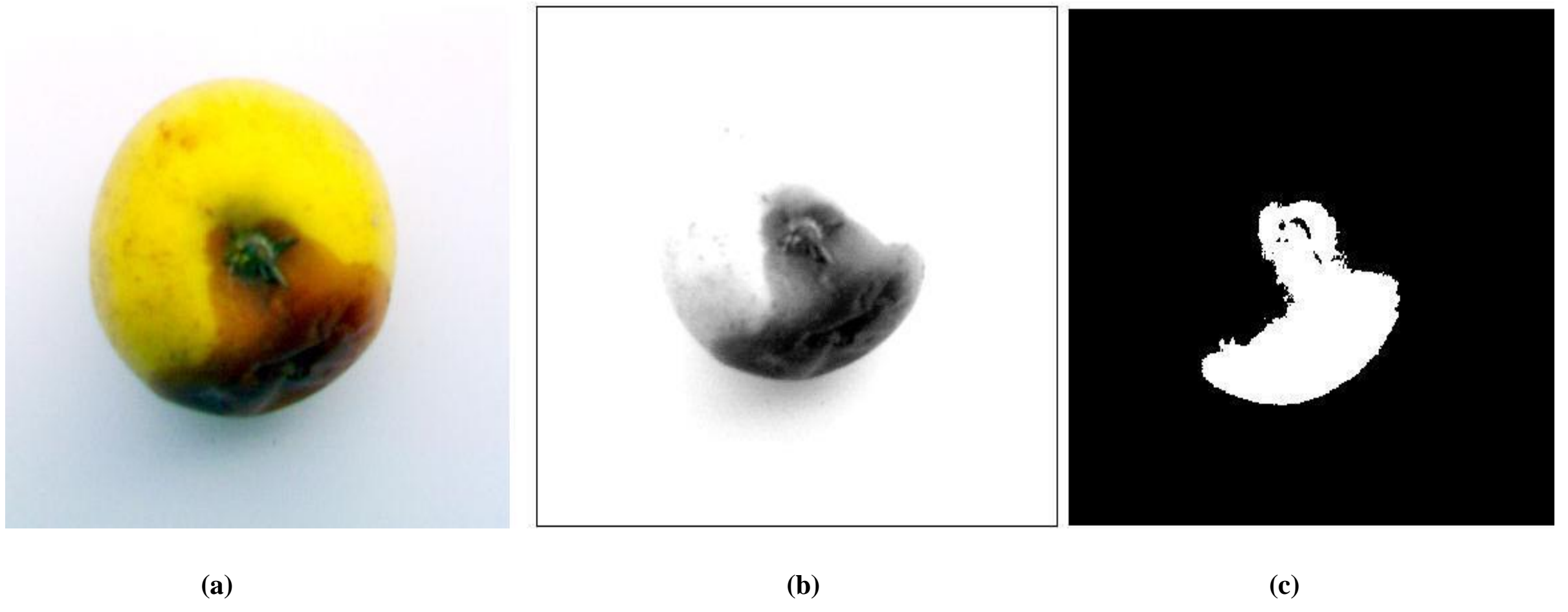

Fig. 8. a) RGB image of a defective golden delicious apple b) Computed gray-scale image c) Segmented image.

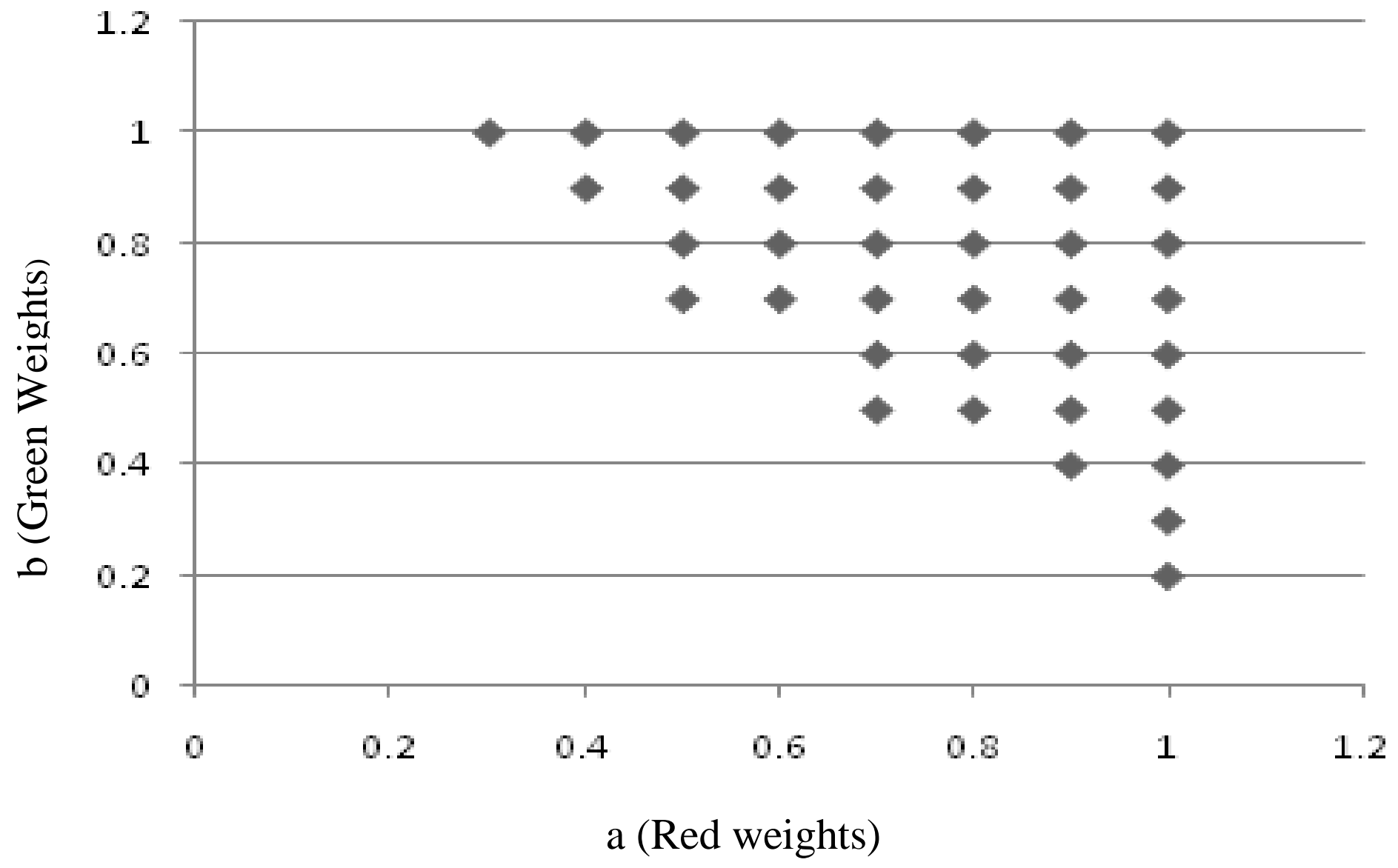

Fig. 9. Distribution of the weights *b* versus *a*.

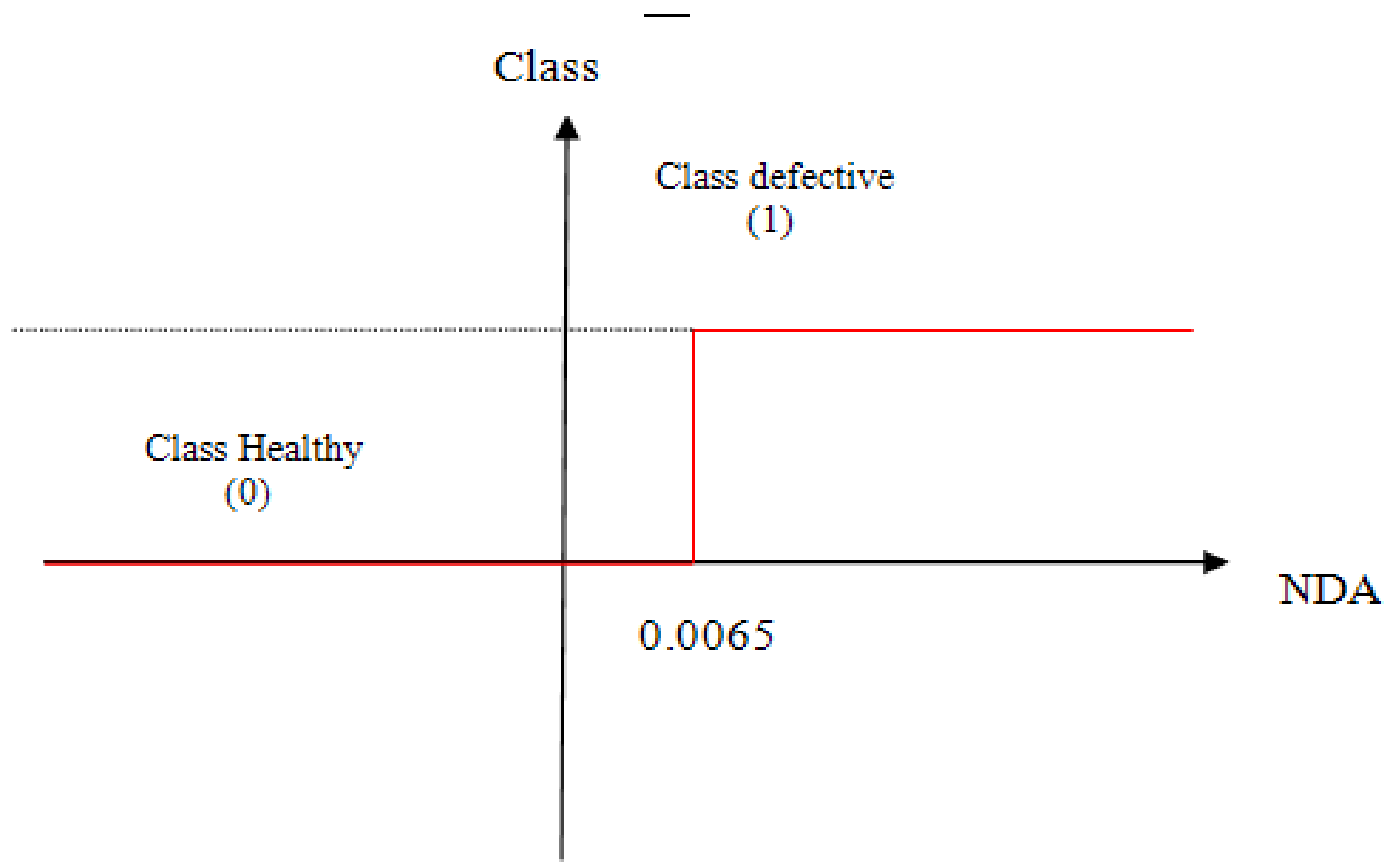

Fig. 10. Schematic of the threshold-based classifier used in this study.

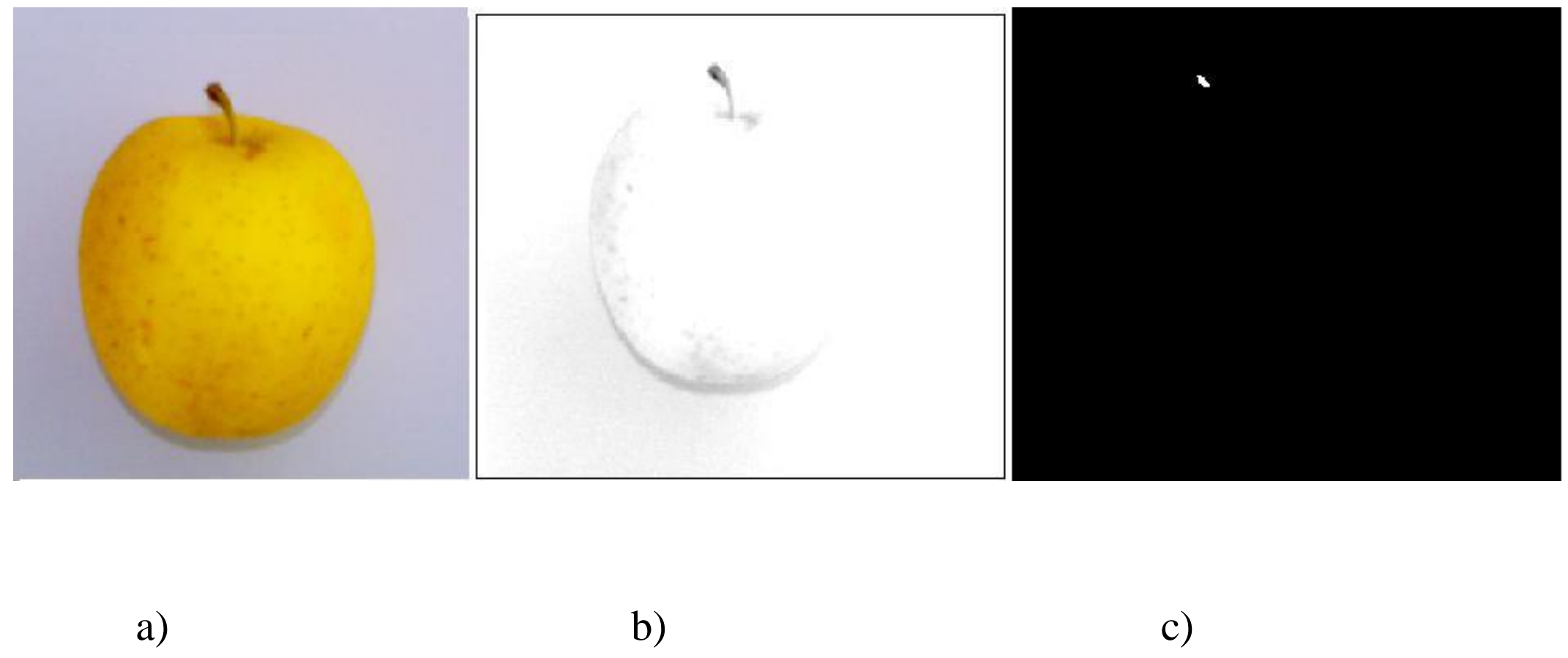

Fig. 11. a) RGB image of healthy golden delicious apple

b) Computed gray scale image using eq. 4. c) Segmented image.